# Discriminative Feature Learning through Feature Distance Loss

Tobias Schlagenhauf
*wbk Institute of Production Science*
*Karlsruhe Institute of Technology*
Karlsruhe, Germany
tobias.schlagenhauf@kit.edu

Yiwen Lin
*wbk Institute of Production Science*
*Karlsruhe Institute of Technology*
Karlsruhe, Germany

Benjamin Noack
*Institute for Intelligent Cooperating Systems*
*Otto von Guericke University*
Magdeburg, Germany

*Abstract*—Ensembles of Convolutional neural networks have shown remarkable results in learning discriminative semantic features for image classification tasks. Though, the models in the ensemble often concentrate on similar regions in images. This work proposes a novel method that forces a set of base models to learn different features for a classification task. These models are combined in an ensemble to make a collective classification. The key finding is that by forcing the models to concentrate on different features, the classification accuracy is increased. To learn different feature concepts, a so-called feature distance loss is implemented on the feature maps. The experiments on benchmark convolutional neural networks (VGG16, ResNet, AlexNet), popular datasets (Cifar10, Cifar100, miniImageNet, NEU, BSD, TEX), and different training samples (3, 5, 10, 20, 50, 100 per class) show the effectiveness of the proposed feature loss. The proposed method outperforms classical ensemble versions of the base models. The Class Activation Maps explicitly prove the ability to learn different feature concepts. The code is available at: https://github.com/2Obe/Feature-Distance-Loss.git

## I. Introduction

Deep convolutional neural networks (CNNs) have recently enabled considerable breakthroughs in many computer vision tasks. It has been proved that CNNs with sufficient depth can achieve remarkable performance in large-scale image recognition tasks because they can extract more complex and comprehensive semantic feature concepts from the images [1][2][3][4]. The convolutional units act as visual concept detectors due to learning to recognize various targets in this process. They can automatically learn discriminative semantic features and locate specific parts of the image responsible for the classification (see Fig. 1). As a result, a convolutional neural network can significantly outperform the best traditional machine learning algorithm, which is based on hand-crafted features[2].

Ensemble learning is a powerful technique for machine learning algorithms, which can achieve excellent performance in various approaches. Therefore, ensemble learning is also widely used in deep learning models [5][6][7][8], and large ensembles of models achieve the best possible results on a task. In general, a range of individual learners is combined with appropriate strategies to enhance the final performance, and the ensemble model has better generalization and discrimination ability. To further improve the performance of the ensemble model, the base models should have considerable diversity. The best ensemble is a set of models that are as different as possible while having as much discriminative power as possible [9].

It is also depicted in Fig. 1 that a single CNN can hardly learn comprehensive semantic features. The CNN tends to concentrate on the most discriminative feature according to its learning capacity, while the model ignores some other valuable features. Therefore, we propose ensemble learning to integrate various complementary semantic features from multiple models. However, different CNNs are probable to learn similar feature concepts if the ensemble models are not further specified. This is akin to a team of experts where each expert shares the same skill set. Homogeneous models make ensemble learning largely pointless [9].

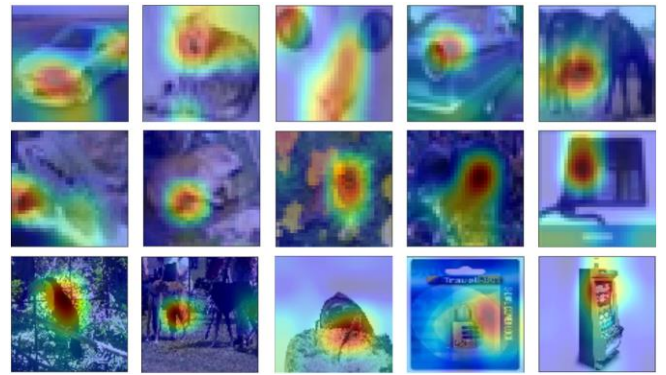

Fig. 1. Gradient-weighted Class Activation Mapping (Grad-CAM) of Cifar10 (top), Cifar100 (middle), and miniImageNet (bottom).

Thus, we propose a novel method to force base models to learn various discriminative features in an image to evaluate a situation better. In our approach, a feature stance loss (further: distance loss) is implemented to quantify the difference between feature concepts learned by base models. On the one hand, we construct feature representation from each model, which can represent the semantic features extracted from an image. On the other hand, we design a distance function to measure the difference between semantic features embedded in the feature representations. After training each base model, a feature fusion model is proposed to integrate the feature information from all base models to make predictions. As a result, the hypothesis is that the combination of adapted base models can achieve better classification performance than an ensemble of models where no care has been taken to learn different features.

To examine this issue, we test our method under various conditions, including different datasets, different dataset sizes, and different CNNs. Our main achievements are 1. We provide a novel distance loss to force CNNs to learn different features. 2. We construct a framework to train and integrate base models with distance loss. 3. We show the effectiveness and generalization ability of our method and evaluate it in both numerical and visual forms. These results also indicate our method's advantageous intervals.

## II. RELATED WORK

### A. Problem Setup

In the here proposed approach the authors want to address the issue that an ensemble of unrestricted base models learns similar image features for classification. The hypothesis which is based on that, is that an ensemble of models will produce better results if they are explicitly forced to learn different features for classification. In order to examine these hypotheses, it is first necessary to understand the state of the art in the field of interpretability of CNN model decisions in the image domain. Based on this, approaches which form CNN ensembles for classification are to be examined. In order to investigate the proposed distance function in the image feature space, an overview of common distance functions in the context of CNN will be given next. The mentioned points are further described in detail.

### B. CNN Interpretability

Image Processing has been the most successful application of deep learning algorithms, and CNNs have been developed a lot in competitions like ImageNet Large Scale Visual Recognition Challenge (ILSVRC). LeNet-5 [1] is one of the most classical convolutional neural networks designed for handwritten digit recognition and is regarded as one of the most representative examples of early CNNs. AlexNet [2] implements a deep convolutional neural network structure on a large-scale image dataset for the first time and shows the absolute predominance of deep learning models. VGG [3] and ResNet [4] make it possible to train very deep CNNs and show excellent performance in recognizing images. The stacking of convolutional layers has been proved to be a powerful method to extract more complex discriminative features for recognition. However, while CNNs can achieve remarkable performance at many vision tasks, it is not easy to understand the nature of the learned representation and why it works so well [10]. They have been treated as black boxes for a long time, and their interpretability is limited because of automatic feature extraction by convolutional layers. Therefore, many pieces of research have been developed to understand CNNs.

All potential semantic features in CNNs can be classified into six types: objects, parts, scenes, textures, material, and colors [11]. Objects and parts can generally be regarded as part patterns, while other semantics belong to textural patterns without explicit shapes. These semantic features emerge in the intermediate convolutional layers and vary through the layers. While beginning layers extract basic features like lines, borders, and corners, deeper layers exhibit high-level features, such as object parts, which are more target-relevant. For example, part detectors emerge in object classifiers [12], and object detectors emerge in scene classifiers [10]. This indicates that CNNs decompose the target of the classification task into multiple lower-level concepts in an interpretable way, such as an object with many parts and a scene with a set of objects. Therefore, it is beneficial that CNNs learn more semantic feature detectors for recognition tasks [11].

For an explicit representation, many methods are used to interpret CNNs. Visualization of filters and feature maps is commonly used to explain the semantic features in CNNs [9]. It explicitly converts the intermediate results to images. Moreover, [13] introduces a deconvolutional network to present pixel-wise responsible patterns, and [14] uses fully connected layers to generate deep convolutional features. [11] also implements Network Dissection to quantify the interpretability of different networks and datasets in numerical values. Moreover, Class Activation Mapping (CAM) [15][16] is another method to visualize semantic features, using the weighted linear sum of the presence of the visual patterns at different spatial locations. A heat map highlights the discriminative regions that are most important to the classification. Furthermore, it has been proved in [15][16][17][18] that CNNs can retain location information of semantic features through the layers and can be used in object localization without any location annotation. This characteristic makes the semantic features extracted by CNNs more interpretable. However, all these methods are used to understand CNNs, and none of them is used for further applications.

### C. Ensemble Learning

Ensemble learning aims to integrate multiple base models to achieve better performance, and there are a wide variety of ensemble methods for machine learning. The voting ensemble method [5] is a very commonly used one and can be implemented in different kinds of base models, such as traditional machine learning models [5] and deep learning models [6]. [7] uses bagging to train deep learning models in parallel and integrate them for classification. Stacking is another prevalent ensemble method, which trains a meta-learner to best combine base models. [8] presents a novel ensemble of deep learning models based on stacking.

For convolutional neural networks, the feature fusion methods are mostly proposed in the image recognition task, which integrates the information at the semantic feature level. In this way, the diversity of features extracted from images is augmented for the recognition task. [19][20][21] introduce methods to fuse features from multiple layers inside a convolutional neural network, and enhance the global features for the recognition task. [22] uses a depthwise convolutional and pointwise convolutional layer to process the fused semantic features to distillate information. At the same time, [23] presents a two-stream CNN to integrate features extracted from two inputs. Moreover, another strategy is handling the data with multiple sizes and providing features with various scales in the CNN models to enrich the feature diversity [24]. The most similar work may be presented in [25], which implements a training strategy to make two subnetworks to learn complementary features. The two-stream features are then fused for the overall classification. However, the two networks are not explicitly forced to learn different features, and possible similar features limit the improvement. There is also no explicit evidence that the two networks learn various complementary semantic features. Our method integrates several convolutional neural networks, which extract different semantic feature concepts from the same image to collect variant-rich features for classification.

*D. Distance Function*

As semantic features can characterize the CNN model, many works try to distinguish different image classes by quantifying the difference between semantic features learned by the CNNs. For example, cosine distance is used to calculate the similarity between different images in the feature space of the Siamese network [26], and the similarity values are used for classification. In addition, [27] presents a method to make predictions only with convolutional layers based on cosine similarity between feature maps. Then, Euclidean distance can measure the content difference in feature maps of different images [28]. Moreover, Structural Similarity (SSIM) and Peak Signal to Noise Ratio (PSNR) are used to compare every two feature maps of a layer, and the similarity is an indicator to prune filters [29]. Finally, in [30], different distance functions, such as Cityblock distance, Minkowski distance, cosine distance, Euclidean distance, and correlation distance, are investigated to measure feature similarity, and cosine distance performs best. These works measure the difference between semantic features inside the CNN, where feature encoding is the same in the convolutional units. In contrast, our method introduces a novel method to compare the feature difference across different CNNs.

In conclusion, CNNs show a remarkable ability to extract semantic features from images, and many works try to enhance the discriminative power of semantic features to improve recognition performance. Most of them fuse features from multiple layers, and others use ensemble methods to integrate multiple models. However, the feature information extracted from base models is not guaranteed to be different, and therefore homogeneous models limit the performance of ensemble learning. Our method uses distance loss to force base models to learn different features from an image.

III. APPROACH

In this section, we propose our distance loss in three main components. The first component is the appropriate global feature representation, and the second component is the distance function. Finally, the training strategy is presented to implement distance loss and fuse various semantic features for classification.

*A. Global Feature Representation*

The distance loss consists of the semantic feature representation and distance function. Before implementing the distance function, we first construct feature representations to interpret which semantic features the base CNN model has learned. This step is the basis for calculating the distance loss, which is described in the next section.

The activation output of the convolutional layers is widely used to interpret semantic features learned from images [10][11][14][15], which are also called feature maps. The feature maps are sparse and distributed semantic feature representations. All semantic concepts are encoded in the distributed convolutional units, and there is a many-to-many relationship between feature concepts and convolutional units [31]. The alignment of disentangled feature representations with convolutional units in a layer varies from CNN to CNN, even with the same architectures [11]. Therefore, we cannot directly compare feature vectors or feature maps as the feature representations across different CNNs. Moreover, one single feature map carries limited semantic information, which is not always meaningful. Only if many feature maps activate the same region can this region be considered to contain practical semantic concepts [17].

Our method uses a simple way to integrate the feature information embedded in the feature maps as the global feature representation. As shown in Fig. 2, the feature maps are summed up pointwise through the channel direction, resulting in an aggregation map. As a result, the feature maps with the shape of $h \times w \times d$ (where $h$ is the height of the feature map, $w$ is the width, and $d$ is the channel number) are integrated into the aggregation map with the shape of $h \times w$. Then, we can ignore the different feature concept distributions in convolutional units across the CNNs and retain spatial information of semantic features.

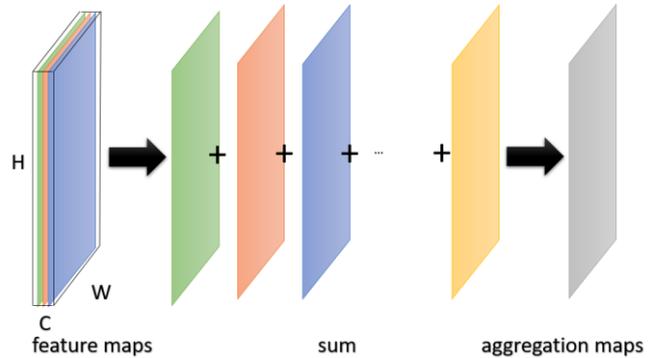

Fig. 2. Integration of the feature maps through channel direction.

We introduce a mask to remove noises and weak semantic features to refine the aggregation map. A threshold $\tau$ is implemented, and all pixel values above this threshold are kept, while other values are set to zero.

$$\tilde{A}(x,y) = \begin{cases} A(x,y) & if\ A(x,y) > \tau \\ 0 & otherwise \end{cases} \quad (1)$$

In (1), $\tilde{A}(x,y)$ refers to the value at position $(x,y)$ in the masked aggregation map, and $A(x,y)$ refers to the value at position $(x,y)$ in the aggregation map. A threshold based on the mean value of the aggregation map is used in our method, i.e., $\tau = mean(A)$. This dynamic threshold can adapt to different aggregation maps. As a result, the most discriminative semantic feature concepts are used to calculate the difference between feature representations across models, which reduces the risk of forcing all base CNN models to learn features on the margin. Otherwise, the base model's performance can be harmed.

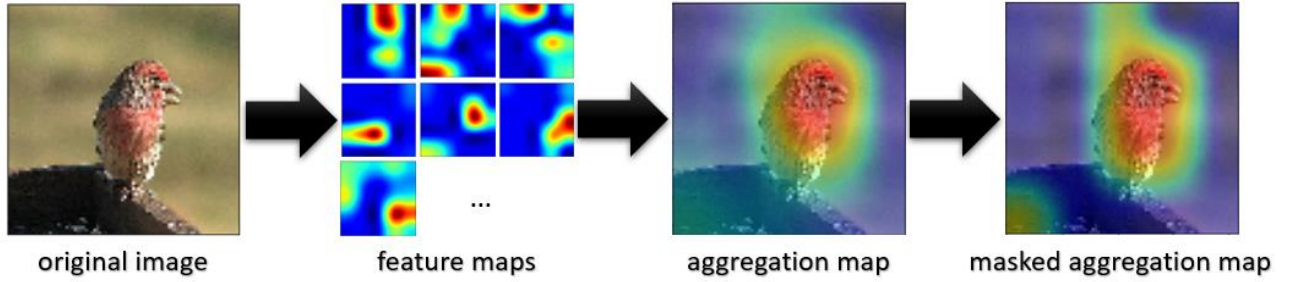

Fig. 3. Pipeline to extract global feature representation for one of the base CNN models.

In addition, convolutional units of higher layers extract more meaningful semantic features, which show excellent discrimination and generalization ability [11]. Therefore, we only extract feature maps from the last convolutional layer of each CNN model. Then, the masked aggregation maps are generated as the global feature representations from base CNN models, respectively, which are used to quantify the difference in semantic features between models in the next section. The process of generating global feature representation for a base CNN model in the ensemble model is depicted in Fig. 3.

*B. Distance Function*

We use a distance function to quantify the difference between semantic features embedded in the global feature representations of different base CNN models. The distance function is based on the combination of cosine and Euclidean distance. On the one hand, cosine distance [26][27][30] can efficiently measure the similarity between two feature vectors regardless of high dimensions, and reflects the relative difference in the direction of the vectors. Therefore, cosine distance pays more attention to the locations of the feature concepts. On the other hand, Euclidean distance interprets the content difference between global feature representations [28]. Unlike cosine distance, Euclidean distance presents the absolute difference in numerical values and works like spatial attention [32][33], which increases the activation level of the critical feature concepts. As a result, the CNN models learn different semantic features in the feature space, and each CNN model also activates its important feature concepts as much as possible. The effectiveness of these two parts is investigated in the ablation study.

As the optimizer is constantly reducing the loss value, and we need to increase the difference between feature representations, the distance loss between any two base models $dloss_{i,j}$ is modified in (2). $v_i$ and $v_j$ refer to two vectorized global feature representations from different global feature representations, while $\alpha$ and $\beta$ are the weights of two distance functions.

$$dloss_{i,j} = \alpha * \frac{v_i^T v_j}{\|v_i\| * \|v_j\|} + \beta * exp(-\|v_i - v_j\|^2) \quad (2)$$

There are two main parts to our distance loss. The first part is cosine similarity, with a limited value between zero and one because all values in the global feature representations are positive. The value zero means very different between these feature representations, and one refers to very similar or the same. The second part is exponential Euclidean distance, and the minus operation ensures that the optimizer can reduce the loss in the direction to increase the difference. In addition, the gradient vanishes with the decrease of the value because of the exponential operator. It is more difficult for the optimizer to reduce the value when it is already small. Therefore, the whole distance loss is dynamically constrained and cannot be minimized such that all models are forced to learn meaningless features on the margins.

*C. Training Strategy*

Our training strategy aims to implement the distance loss for training base CNN models and integrate the feature information in the ensemble model for classification.

We propose the joint training for five base CNN models with the same architecture, including input size, layer structure, and output size. As shown in Fig. 4, every base model is trained individually to perform classification with the same training samples. As the multi-class classification task is tested in our method, the softmax activation function and cross entropy loss are implemented for every classifier. On the other hand, the feature maps at the last convolutional layers are extracted from the CNN models. The global feature representations are generated to present the semantic features learned by the base CNN models. The distance function is then used to calculate the distance loss.

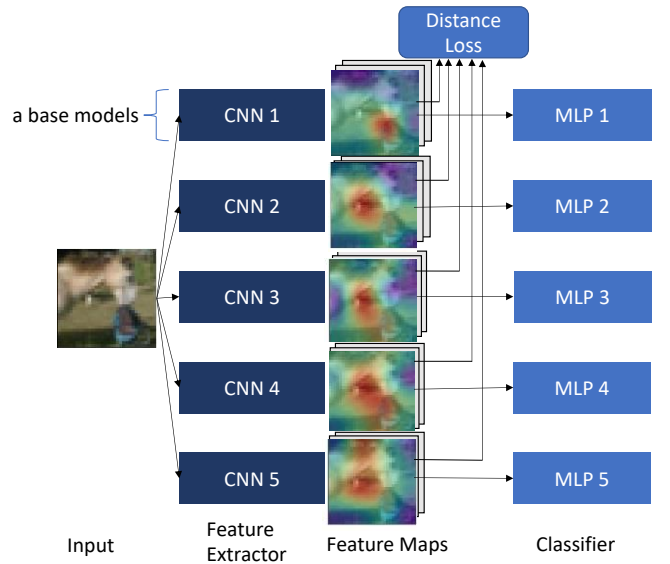

Fig. 4. The main framework for training the base models.

The whole training loss consists of classification loss and distance loss. In (3), the first part is cross entropy loss for the classification, where $y_k^i$ is the true label of the kth class of the training sample, and $\hat{y}_k^i$ refers to the predicted probability of the kth class in the ith base model.

$$loss = \sum_{i=1}^{m}(-\sum_{k=1}^{n} y_k^i \log \hat{y}_k^i) + \sum_{i,j,i\neq j} dloss_{i,j} \quad (3)$$

Besides, $m$ is the number of base models. i.e., five, and $n$ is the class number, which varies between different datasets. The second part is the distance loss, which will be calculated between every two different base models.

After training, all base models are integrated into an ensemble model, maximizing the benefits of high feature diversity and improving performance. In our method, we propose one of the feature fusion models and make an ensemble of the base models at the semantic feature level. Unlike many traditional ensemble methods that use the entire base models, we only use the convolutional part of the CNN base models as feature extractors and concatenate the feature maps in the direction of the channel. Then, the fused feature maps are fed into a single new classifier for classification. In other words, we use the trained CNN models to construct a whole end-to-end model at last. The base CNN models are frozen, and the new fully connected layers with softmax activation and categorical cross entropy loss are trained for classification. The final classification model is a fully connected model with one hidden layser with relu activation and 128 nodes. The hidden layer is followed by a dropout layer with 0.5 dropout and an output layer with softmax activation. The final model is trained for 50 epochs with batch-size of 10 and a learning rate of 10-4. Therefore, all semantic features from the base CNN models are processed together, increasing feature diversity for classification (Fig. 4).

## IV. EXPERIMENTS & RESULTS

In this section, we first introduce the datasets and the implementation details of the experiments. Then, we present our method's performance under different conditions to show its effectiveness and generalization ability. Finally, the initialization strategies and effectiveness of different distance functions are explored in the ablation study. The experiments are also made to find the best ensemble methods to integrate the semantic features.

### A. Datasets and Implementation Details

We conduct our experiments on six datasets, including Cifar10 [34], Cifar100 [34], miniImageNet [35], NEU [36], TEX [37], and BSD [38]. As shown in Fig. 5, Cifar10 and Cifar100 are well-known object classification datasets, and each has 60000 32×32 color images. There are ten classes for Cifar10 and 100 classes for Cifar100. MiniImageNet has a higher complexity due to the use of original ImageNet, but requires much less resources, making it convenient for rapid prototyping and experimentation. There are 100 classes with 600 84×84 color images each. In addition, we also introduce three technical datasets, which are different from the object-based datasets. The NEU dataset is based on the metallic surface defect and has 1800 200×200 grayscale images for six classes. The TEX dataset (originally called fabric dataset) shows five different types of failures in textiles and one good class. Each class has 18000 64×64 grayscale images, and there are 108000 samples. The last dataset is BSD, showing failures on ball screw drives. This dataset has 21835 150×150 color images, and all images are labeled with two classes, i.e., defect and no defect, which are roughly equally divided. As a result, we can evaluate our method on different datasets, including object-based and nonobject-based datasets with different levels of semantic features. Generally, we split these datasets randomly in 60% for training samples, 20% for validation samples, and 20% for testing samples.

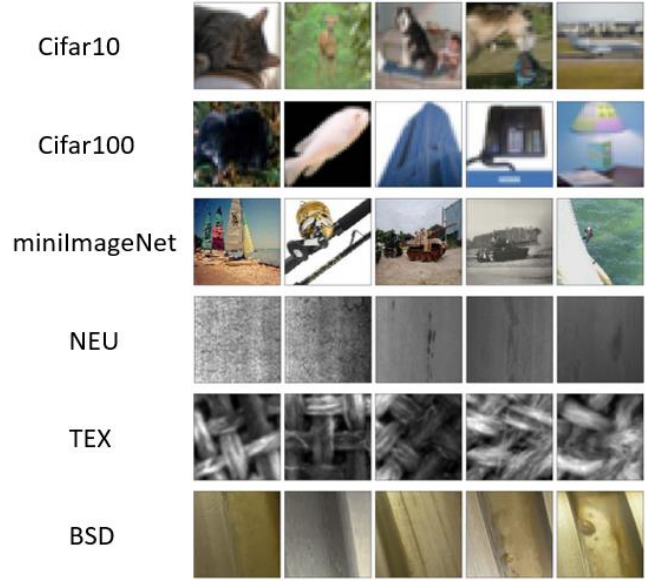

Fig. 5. The images of Cifar10, Cifar100, miniImageNet, NEU, TEX, and BSD.

In addition to different datasets, the base models in our experiments are based on famous CNN architectures, including VGG16 [3], ResNet12 [4][39], and AlexNet [2]. The five base models implement the same CNN architectures, which are initialized by the He normal initializer with five different random seeds, such as 1, 2, 3, 4, and 5. Therefore, we can have relatively stable, but various initial states and the effectiveness is investigated in the ablation study.

In the training phase, the base models are trained jointly for 300 epochs with a learning rate of $10^{-4}$, and all training samples are used once at each epoch. Meanwhile, we save the entire model with the best performance during the training phase, where the average accuracy of the base models is an indicator. Besides, all images are randomly transformed by combining image augmentation methods, including rotation, horizontal and vertical flipping, Laplace noise, and translation, while training to generate different training samples as many as possible. For evaluation, we use unseen testing samples to calculate the classification accuracy and generate CAMs to visualize the semantic features learned by the base models.

### B. Results

Using the proposed components, we test our method under various conditions, including different datasets (Cifar10, Cifar100, miniImageNet, NEU, TEX, and BSD), numbers of training samples (3, 5, 10, 20, 50, 100, and 400), and CNN architectures (VGG, ResNet, and AlexNet). On the one hand, the classification accuracy of the ensemble model with and without distance loss is listed in tables as numerical results. On the other hand, CAMs are generated in figures to visualize semantic features learned by the base models.

The results for Cifar10, Cifar100, and miniImageNet are shown in Tab. I. It is obvious that the ensemble model with distance loss can consistently outperform the base models and the ensemble model without distance loss. The distance loss always has a positive effect on the performance of the ensemble model. For example, the distance loss can improve the ensemble performance by 3.94% from 49.47% to 53.41% for miniImagesNet with 100 training samples per class in ResNet.

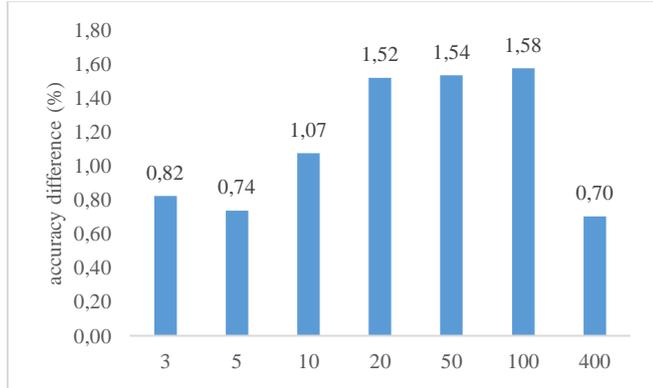

Fig. 6. Effectiveness of the distance loss for various amounts of training samples, including 3, 5, 10, 20, 50, 100, and 400 samples per class. The values are calculated by averaging the accuracy differences (%) between the ensemble model with and without distance loss. Cifar10, Cifar100, miniImageNet, and various CNN architectures are considered.

However, the improvement is not constant for all conditions, and the distance loss has different performances with different dataset sizes. As shown in Fig. 6, the method dominates middle-scale datasets like 100 samples per class. When the number of training samples is too small, it is not accessible to learn precise semantic features, making the global feature representation not interpretable. In contrast, the model can learn more discriminative features with sufficient training samples. As the total discriminative features in an image are fixed, the potential to increase the feature diversity can be reduced in this case.

Besides, the CNN architectures have a significant influence on the performance of the distance loss (see Fig. 9). Generally speaking, the ResNet architecture achieves the largest improvement, while the VGG and AlexNet architectures are in second and third place. The residual block structures in the ResNet architecture make the integration of multi-layer semantic features possible, which can increase the representative ability of the feature maps at the high layers [19]. More discriminative semantic features can be encoded into our global feature representation, making the comparison between semantic features more meaningful..

TABLE I. CLASSIFICATION ACCURACY (%) WITH DIFFERENT NUMBERS OF TRAINING SAMPLES IN VGG, RESNET, AND ALEXNET FOR CIFAR10, CIFAR100, AND MINIIMAGENET.

| Architecture | Dataset size | Base model | No distance loss | Distance loss | Base model | No distance loss | Distance loss | Base model | No distance loss | Distance loss |
|---|---|---|---|---|---|---|---|---|---|---|
| | | *Cifar10* | | | *Cifar100* | | | *miniImageNet* | | |
| VGG | 3 | 22.50 | 23.43 | 24.85 | 8.56 | 10.35 | 11.04 | 6.80 | 8.19 | 8.78 |
| | 5 | 27.31 | 28.89 | 29.73 | 10.31 | 12.44 | 13.05 | 8.94 | 10.51 | 11.47 |
| | 10 | 30.05 | 31.64 | 32.90 | 15.55 | 18.63 | 20.62 | 14.85 | 17.62 | 18.02 |
| | 20 | 37.60 | 41.93 | 43.52 | 22.47 | 26.14 | 29.30 | 21.77 | 25.13 | 26.52 |
| | 50 | 49.93 | 54.04 | 55.35 | 34.84 | 41.00 | 42.94 | 33.21 | 37.58 | 38.93 |
| | 100 | 58.00 | 63.77 | 65.17 | 46.30 | 51.19 | 53.27 | 43.90 | 48.21 | 49.27 |
| | 400 | 74.02 | 78.08 | 78.83 | 62.86 | 67.53 | 68.32 | 63.15 | 67.59 | 68.28 |
| ResNet | 3 | 22.54 | 22.13 | 23.29 | 7.29 | 10.18 | 11.03 | 6.42 | 8.94 | 10.06 |
| | 5 | 26.45 | 28.85 | 29.66 | 9.01 | 12.81 | 13.38 | 7.87 | 12.42 | 13.75 |
| | 10 | 28.45 | 31.48 | 32.95 | 14.19 | 20.42 | 22.18 | 11.85 | 19.32 | 21.00 |
| | 20 | 36.38 | 42.27 | 43.61 | 19.80 | 28.52 | 30.08 | 16.55 | 27.27 | 29.10 |
| | 50 | 45.77 | 53.76 | 55.23 | 32.97 | 43.05 | 44.74 | 28.28 | 38.45 | 42.26 |
| | 100 | 55.63 | 62.75 | 63.78 | 43.56 | 52.54 | 54.16 | 38.17 | 49.47 | 53.41 |
| | 400 | 71.91 | 78.29 | 78.69 | 62.01 | 69.04 | 69.43 | 57.51 | 67.61 | 69.78 |
| AlexNet | 3 | 22.20 | 22.63 | 23.04 | 8.92 | 10.12 | 10.55 | 6.97 | 7.61 | 8.36 |
| | 5 | 26.51 | 26.71 | 27.46 | 10.47 | 12.03 | 12.46 | 8.89 | 10.21 | 10.55 |
| | 10 | 31.67 | 32.72 | 33.10 | 14.68 | 17.56 | 18.09 | 12.20 | 15.73 | 15.93 |
| | 20 | 38.57 | 40.76 | 42.17 | 20.87 | 24.78 | 25.60 | 16.52 | 20.42 | 21.01 |
| | 50 | 47.06 | 50.53 | 51.51 | 30.84 | 36.05 | 36.71 | 24.42 | 30.37 | 30.98 |
| | 100 | 53.90 | 58.94 | 60.51 | 38.94 | 45.63 | 46.57 | 31.48 | 38.08 | 38.62 |
| | 400 | 67.59 | 72.64 | 73.31 | 54.46 | 61.00 | 61.34 | 45.67 | 52.59 | 52.73 |

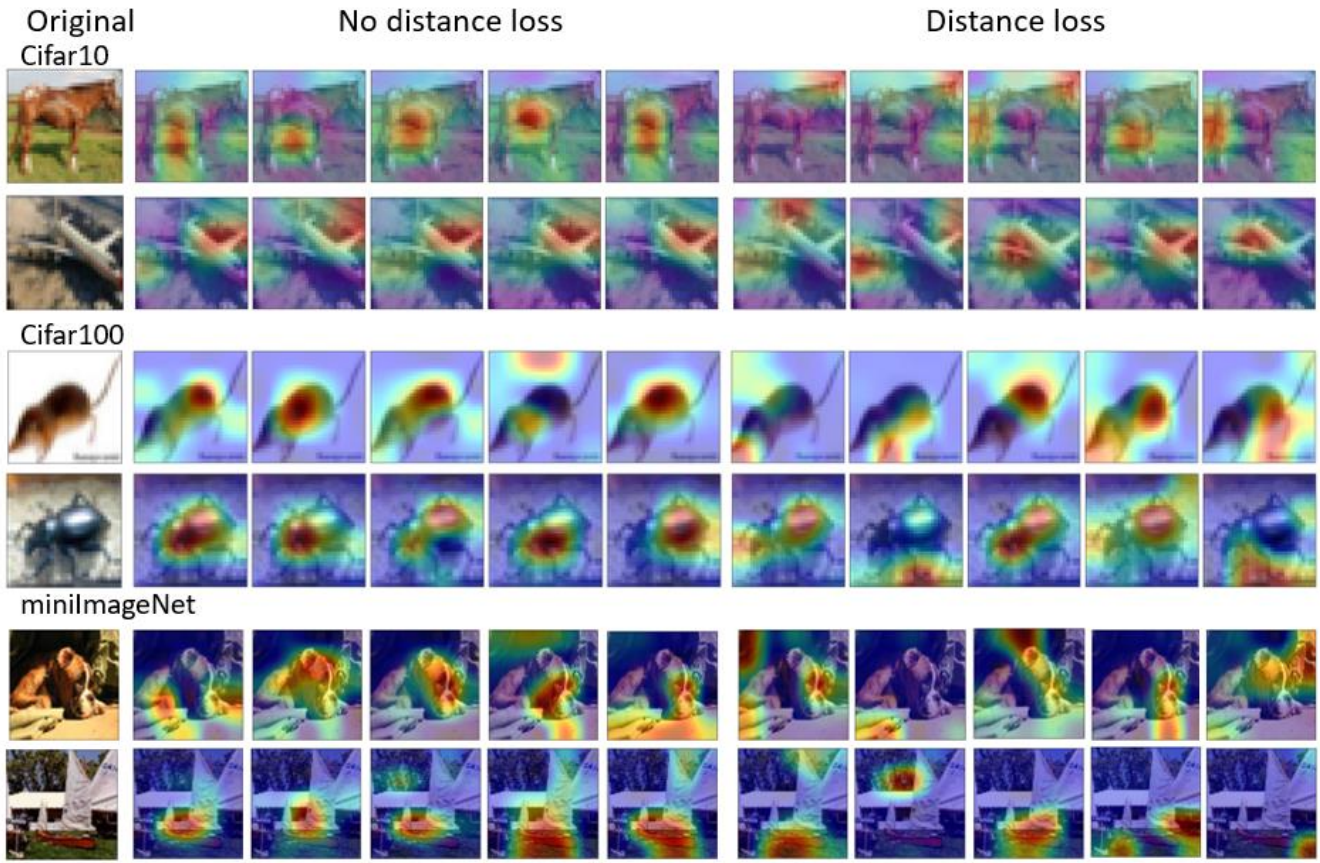

Fig. 7. CAMs of the five base models without distance loss (left) and with distance loss (right) for Cifar10, Cifar100, and miniImageNet.

TABLE II.  CLASSIFICATION ACCURACY (%) WITH DIFFERENT NUMBERS OF TRAINING SAMPLES IN VGG, RESNET, AND ALEXNET FOR NEU, TEX, AND BSD.

| Architecture | Dataset size | Base model | No distance loss | Distance loss | Base model | No distance loss | Distance loss | Base model | No distance loss | Distance loss |
|---|---|---|---|---|---|---|---|---|---|---|
| | | *NEU* | | | *TEX* | | | *BSD* | | |
| VGG | 3 | 44.17 | 43.61 | 44.44 | 21.09 | 26.22 | 28.14 | 69.11 | 63.14 | 73.70 |
| | 5 | 52.78 | 51.94 | 51.94 | 23.43 | 26.43 | 29.53 | 73.09 | 68.47 | 69.89 |
| | 10 | 69.44 | 67.78 | 69.17 | 34.22 | 37.02 | 37.17 | 81.56 | 75.69 | 79.74 |
| | 20 | 77.50 | 71.39 | 78.89 | 39.81 | 41.77 | 41.11 | 88.65 | 84.95 | 83.10 |
| | 50 | 88.89 | 83.06 | 90.28 | 50.96 | 51.37 | 52.61 | 94.30 | 93.17 | 94.06 |
| | 100 | 96.11 | 94.44 | 96.67 | 56.09 | 58.00 | 59.36 | 92.90 | 92.61 | 92.88 |
| ResNet | 3 | 55.19 | 57.78 | 51.94 | 33.67 | 30.90 | 32.14 | 68.91 | 65.79 | 74.51 |
| | 5 | 57.04 | 63.89 | 59.17 | 35.66 | 35.89 | 33.14 | 75.27 | 72.64 | 74.61 |
| | 10 | 70.37 | 64.72 | 71.67 | 41.58 | 41.38 | 41.69 | 80.12 | 80.55 | 83.84 |
| | 20 | 75.14 | 74.44 | 76.39 | 46.17 | 46.19 | 46.87 | 86.99 | 87.25 | 86.25 |
| | 50 | 89.17 | 85.83 | 90.56 | 52.92 | 55.37 | 55.43 | 88.43 | 90.64 | 90.79 |
| | 100 | 95.65 | 96.11 | 96.39 | 56.75 | 58.41 | 58.69 | 91.91 | 91.85 | 91.09 |
| AlexNet | 3 | 44.44 | 44.72 | 47.78 | 32.21 | 32.61 | 33.03 | 73.10 | 76.67 | 77.55 |
| | 5 | 48.52 | 51.94 | 53.33 | 34.42 | 33.70 | 35.71 | 76.46 | 76.99 | 78.83 |
| | 10 | 55.28 | 55.56 | 55.56 | 38.61 | 40.13 | 39.50 | 83.53 | 87.75 | 87.84 |
| | 20 | 61.67 | 60.83 | 60.83 | 41.28 | 42.66 | 42.65 | 85.39 | 87.65 | 90.35 |
| | 50 | 82.31 | 73.89 | 83.89 | 48.70 | 49.61 | 50.06 | 90.63 | 90.10 | 90.32 |
| | 100 | 88.43 | 86.67 | 89.72 | 53.33 | 54.07 | 53.83 | 91.02 | 89.66 | 90.42 |

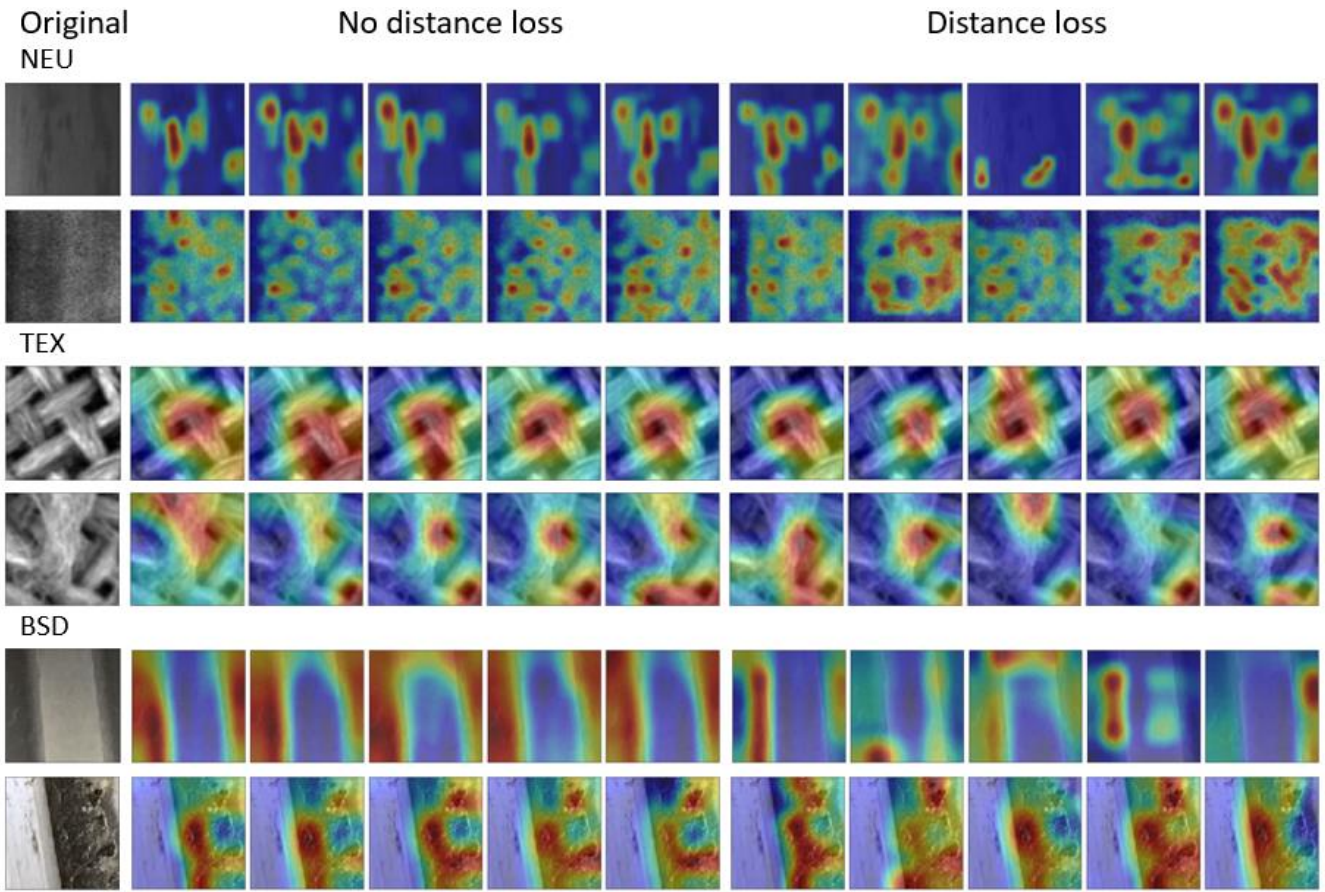

Fig. 8. CAMs of the five base models without distance loss (left) and with distance loss (right) for NEU, TEX, and BSD.

Furthermore, we also present the effectiveness of the distance loss in Fig. 7 for clear visualization. The base CNN models without distance loss tend to concentrate on a relatively constant part of the object in the image, indicating similar semantic feature concepts. In contrast, the base CNN models with distance loss can have various options for semantic feature concepts. The five base CNN models focus on different object parts, i.e., various features. Consequently, it can be concluded that the distance loss can increase the feature diversity in the ensemble model and improve the classification performance compared to the ensemble model without distance loss.

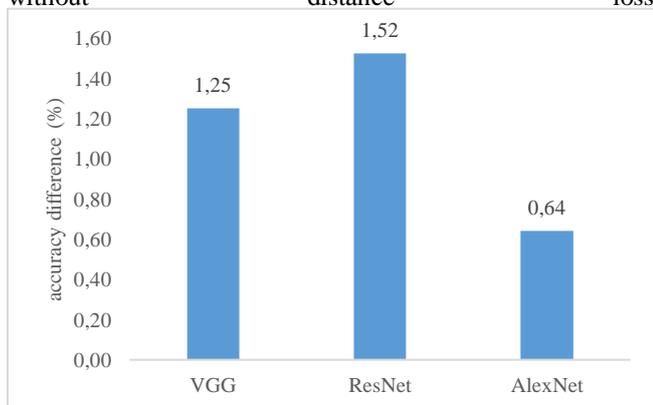

Fig. 9. Effectiveness of the distance loss for VGG, ResNet, and AlexNet architectures. The values are calculated by averaging the accuracy differences (%) between the ensemble model with and without distance loss. Cifar10, Cifar100, miniImageNet, and various training samples are considered.

In addition to object-based datasets, we also test our distance loss on technical datasets, i.e., NEU, TEX, and BSD, which are not based on objects. As depicted in Tab. II, the performance of the distance loss with these technical datasets is different from Cifar10, Cifar100, and miniImageNet. The ensemble model without distance loss cannot steadily achieve better performance than the ensemble model without distance loss or the base models. In contrast, the base models can also achieve an equivalent performance to the ensemble model with distance loss. There is no explicit tendency for the performance of the distance loss, and therefore, the distance loss does not work well on these technical datasets for classification.

On the other hand, the visualization of the semantic features is shown in Fig. 8, and we can see that the base models with distance loss can also focus on the similar semantic features in the images. The base models can even be forced to learn nothing discriminative or much fewer features because of the distance loss. In contrast, every base model can already learn comprehensive features for classification. As a result, we cannot increase the feature diversity by the ensemble model with distance loss, and the distance loss can even reduce the classification performance under some conditions.

As shown in Fig. 10, the ensemble model with distance loss on Cifar10, Cifar100, and miniImageNet achieves much better performance than on NEU, TEX, and BSD. On the one hand, the technical datasets are based on textural patterns without explicit shapes or locations, such as lines, corners, and colors. These low-level features are simple and likely to be shared by different classes at lower layers [40], making the feature representations less meaningful. Therefore, the discriminative feature concepts in the datasets like NEU, TEX, and BSD are limited, and redundant feature information leads to overfitting in the ensemble model. On the other hand, object-based datasets like Cifar10, Cifar100, and miniImageNet have much more part-level or object-level features. These high-level features are more interpretable and class-specific [40], making the feature representation more representative. As a result, it is concluded that the ensemble model with distance loss can improve the classification performance on the datasets, which are rich in semantic features. If there is only one or very few important features, forcing the models to learn different features is misleading.

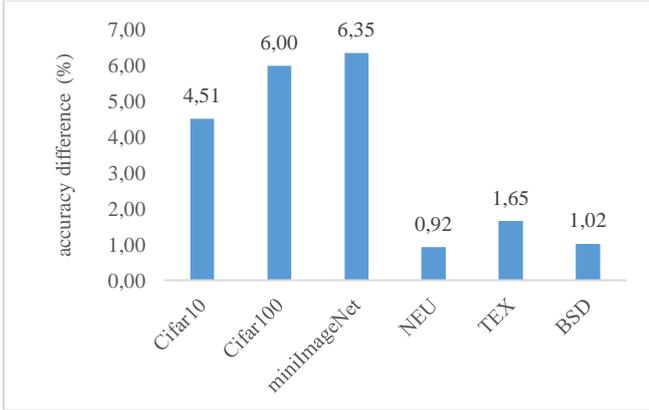

Fig. 10. Average improvement of the ensemble model with distance loss compared to the base models. Different CNN architectures and dataset sizes are considered for each dataset.

## C. Ablation Study

In order to study the effectiveness of the components in our method, we make this ablation study. It aims to find the best components and improve the model's performance as much as possible. Therefore, we set the baseline for the ablation study: 100 samples per class of the Cifar10 dataset in the VGG16 models, providing a stable and comparable condition, and the classification accuracy of the ensemble model is presented as the final result.

### 1) Initialization Strategy

TABLE III. CLASSIFICATION ACCURACY (%) USING DIFFERENT INITIALIZATION STRATEGIES.

|  | **Initialization strategy** | **Final result** |
|---|---|---|
| No distance loss | None | 63.24 |
|  | Same | 63.34 |
|  | Different | 63.49 |
| Distance loss | None | 64.30 |
|  | Same | 64.07 |
|  | Different | 65.17 |

We introduce three strategies to initialize the five base models. The first one is initialization without explicit random seed (None), resulting in utterly random initialization. Then, we initialize the five base models with random seeds, i.e., 1, 2, 3, 4, and 5, and it is guaranteed that the base models have different start states. At last, the third strategy is initializing the base models with the same random seed. The accuracy averages the results of five tests, using random seeds 1, 2, 3, 4, and 5, respectively. As shown in Tab. III, the three initialization strategies are implemented in our framework with and without distance loss. Although the accuracy can only be slightly affected by the initialization strategy, the distance loss with the initialization strategy of different random seeds can achieve much better performance. Obviously, the different start states of the base models help our framework with distance loss converge to a better solution during the training. Our method implements this initialization strategy for all experiments

### 2) Distance Function

We test various distance functions in the distance loss, which are widely used for feature comparison, including cosine distance, Euclidean distance, SSIM, and our Consine&Euclidean distance function. In addition, the appropriate weight for each distance loss is also investigated.

TABLE IV. CLASSIFICATION ACCURACY (%) USING DIFFERENT DISTANCE FUNCTIONS AND WEIGHTS.

| **Distance function** | **Loss weight** | **Final result** |
|---|---|---|
| None | None | 63.49 |
| Cosine [26] | 0.1 | 63.94 |
|  | 1 | 63.79 |
|  | 2 | 63.49 |
| Euclidean [28] | 1 | 64.25 |
|  | 10 | 64.04 |
|  | 20 | 63.78 |
| SSIM [29] | 1 | 64.27 |
|  | 5 | 64.66 |
|  | 10 | 64.98 |
|  | 20 | 64.77 |
| Cosine + Euclidean distance | 0.2 + 1 | 63.95 |
|  | 0.5 + 10 | 63.60 |
|  | 1 + 10 | 65.17 |
|  | 1 + 20 | 64.23 |

In Tab. IV, our Cosine & Euclidean distance function with the loss weight of 1 & 10 achieves the best accuracy. It is also interesting that the performance increases with increasing loss weight at the beginning but decreases later. The reason is that the larger loss weight can force the base models to focus on more different semantic features and increase the feature diversity for classification. However, the base models are proposed to learn features in the background if the loss weight is too large, which leads to poor classification performance of the base models.

### 3) Ensemble Method

TABLE V. CLASSIFICATION ACCURACY (%) USING DIFFERENT ENSEMBLE METHODS.

|  | **Ensemble method** | **Final result** |
|---|---|---|
| No distance loss | No ensemble method | 58.00 |
|  | Hard voting method [5] | 62.42 |
|  | Soft voting method [5] | 63.29 |
|  | Stacking ensemble [8] | 63.52 |
|  | Pooling approach [14] | 59.72 |
|  | Trainable fusion method [22] | 62.68 |
|  | Addition fusion method [20] | 62.56 |
|  | Concatenation fusion method [20] | 63.77 |
| Distance loss | No ensemble method | 58.08 |
|  | Hard voting method [5] | 62.35 |
|  | Soft voting method [5] | 63.01 |
|  | Stacking ensemble [8] | 64.17 |
|  | Pooling approach [14] | 62.05 |
|  | Trainable fusion method [22] | 63.79 |
|  | Addition fusion method [20] | 63.56 |
|  | Concatenation fusion method [20] | 65.17 |

Last, the authors also investigate the effect of different ensemble methods from related works. All ensemble methods are implemented in our framework with and without distance loss. It is depicted in Tab. V that the concatenation fusion method consistently achieves the best performance in the ensemble model. Interestingly, with or without distance loss, our framework gives similar results, using the voting methods or stacking ensemble method with a similar base model performance. In contrast, the feature fusion models can better use high feature diversity and improve classification performance.

## V. Conclusion

In this work, we proposed a novel method to make more efficient ensemble learning of multiple base CNN models without pretraining and transfer learning. The critical component is a distance loss, which forces the base models to learn different semantic features from images. The distance loss first generates global feature representations from the base models. The semantic features learned by the base model are integrated into the masked aggregation map. Then, we use a distance function to quantify the difference in feature concepts of the global feature representations. This distance function is implemented between every two base models, and the sum is calculated as the distance loss. The experiments show that the distance loss can enhance the feature diversity and increase the classification performance in the ensemble model for the datasets, such as Cifar10, Cifar100, and miniImageNet, which have rich discriminative semantic features.

In the future, we will consider finding ways to automatically determine if it is helpful for the model to use the distance loss with a specific dataset. Although we propose a fixed weight for the distance loss, the best weight varies in different conditions, especially in different datasets. Therefore, a learnable weight parameter is a possible solution, which automatically learns how strong the distance loss should be weighted. Then, the loss weight can adapt to the number of features the dataset contains. Moreover, further study can be made to increase the representative ability of the global feature representation. A good idea comes from the residual structure, which integrates the multi-layer features. We can thus use more powerful CNN architectures, such as DenseNet [41], or directly encode the feature information from multiple layers into the global feature representation. Finally, the number of the base models is also an exciting investigation direction. The heat maps above show that multiple base models with distance loss concentrate on similar parts, indicating redundant information. Some base models can then be pruned to reduce resource consumption and prevent overfitting. Meanwhile, more base models can also be added to enhance the feature diversity in the ensemble model when the dataset contains more discriminative semantic features.

## References

[1] Lecun, Y., Bottou, L., Bengio, Y. & Haffner, P. (1998). Gradient-based learning applied to document recognition. Proceedings of the IEEE, 86(11), 2278–2324. https://doi.org/10.1109/5.726791.

[2] A. Krizhevsky, I. Sutskever, and G. E. Hinton, "ImageNet classification with deep convolutional neural networks," Advances in Neural Information Processing Systems 25 (NIPS 2012).

[3] K. Simonyan and A. Zisserman, "Very deep convolutional networks for large-scale image recognition," ICLR, 2015.

[4] K. He, X. Zhang, S. Ren, J. Sun, "Deep residual learning for image recognition," arXiv:1512.03385v1 [cs.CV] 10 Dec 2015.

[5] J. Brownlee, Ensemble Learning Algorithms With Python: Make Better Prediction with Bagging, Boosting, and Stacking, Machine Learning Mastery, 2021.

[6] M. Turkoglu, D. Hanbay and A. Sengur, "Multi-model LSTM-based convolutional neural networks for detection of apple diseases and pests," Journal of Ambient Intelligence and Humanized Computing, 2019.

[7] Essa, E. & Xie, X. (2021). An Ensemble of Deep Learning-Based Multi-Model for ECG Heartbeats Arrhythmia Classification. IEEE Access, 9, 103452–103464. https://doi.org/10.1109/ACCESS.2021.3098986.

[8] A. Khamparia, A. Singh, D. Anand, D. Gupta, A. Khanna, N. A. Kumar and J. Tan, "A novel deep learning-based multi-model ensemble method for the prediction of neuromuscular disorders," Neural Computing and Applications, 2020.

[9] F. Chollet, Deep Learning with Python, Manning Publications Co., 2018, pp264-267.

[10] B. Zhou, A. Khosla, A. Lapedriza, A. Oliva, A. Torralba, "Object detectors emerge in deep scene CNNs," arXiv:1412.6856v2 [cs.CV] 15 Apr 2015.

[11] D. Bau, B. Zhou, A. Khosla, A. Oliva, A. Torralba, "Network dissection: quantifying interpretability of deep visual representations," arXiv:1704.05796v1 [cs.CV] 19 Apr 2017.

[12] A. Gonzalez-Garcia, D. Modolo, V. Ferrari, "Do semantic parts emerge in covolutional neural networks?" arXiv:1607.03738v5 [cs.CV] 20 Sep 2017.

[13] M. D Zeiler and R. Fergus, "Visualizing and understanding convolutional networks," arXiv:1311.2901v3 [cs.CV] 28 Nov 2013.

[14] X. Wei, C. Xie and J. Wu, "Mask-CNN: localization parts and selecting descriptors for fine-grained image recognition," arXiv:1605.06878v1 [cs.CV] 23 May 2016.

[15] B. Zhou, A. Khosla, A. Lapedriza, A. Oliva and A. Torralba, "Learning deep features for discriminative location," arXiv:1512.04150v1 [cs.CV] 14 Dec 2015.

[16] R. R. Selvaraju, M. Cogswell, A. Das, R. Vedantam, D. Parikh and D. Batra, "Grad-CAM: visual explanations from deep networks via gradient-based localization," arXiv:1610.02391v4 [cs.CV] 3 Dec 2019.

[17] X. Wei, J. Luo, J. Wu and Z. Zhou, " Selective convolutional descriptor aggregation for fine-grained image retrieval," arXiv:1604.04994v2 [cs.CV] 13 Jul 2017.

[18] F. Chen, G. Huang, J. Lan, Y. Wu, C. Pun, W. Ling and L. Cheng, "Weakly supervised fine-grained image classification via salient region localization and different layer feature fusion," Appl. Sci. 2020, 10(13), 4652.

[19] Y. Hua, L. Mou and X. Zhu, "LAHNet: a convolutional neural network fusion low- and high-level features for aerial scene classification," IEEE, 2018.

[20] Ma, C., Mu, X. & Sha, D. (2019). Multi-Layers Feature Fusion of Convolutional Neural Network for Scene Classification of Remote Sensing. IEEE Access, 7, 121685–121694. https://doi.org/10.1109/ACCESS.2019.2936215.

[21] Y. Liu, M. Cheng, X. Hu, K. Wang and X. Bai, "Richer convolutional features for edge detection," arXiv:1612.02103v3 [cs.CV] 3 Jul 2019.

[22] J. Kim, M. Hyun, I. Chung and N. Kwak, "Feature fusion for online mutual knowledge distillation," arXiv:1904.09058v2 [cs.CV] 21 Jul 2020.

[23] Hu, J., Mou, L., Schmitt, A. & Zhu, X. X. (2017). FusioNet: A two-stream convolutional neural network for urban scene classification using PolSAR and hyperspectral data. In 2017 Joint Urban Remote Sensing Event (JURSE) (S. 1–4). IEEE. https://doi.org/10.1109/JURSE.2017.7924565.

[24] Li, E., Xia, J., Du, P., Lin, C. & Samat, A. (2017). Integrating Multilayer Features of Convolutional Neural Networks for Remote Sensing Scene Classification. IEEE Transactions on Geoscience and Remote Sensing, 55(10), 5653–5665. https://doi.org/10.1109/TGRS.2017.2711275.

[25] Hou, S., Liu, X. & Wang, Z. (2017). DualNet: Learn Complementary Features for Image Recognition. In 2017 IEEE International Conference on Computer Vision (ICCV) (S. 502–510). IEEE. https://doi.org/10.1109/ICCV.2017.62.

[26] T. Schlagenhauf, F. Yildirim and B. Brueckner, " Siamese basis function networks for data-efficient defect classification in technical domains," arXiv:2012.01338, 2021.

[27] K. Park and D. Kim, "Accelerating image classification using feature map similarity in convolutional neural networks," Appl. Sci. 2019, 9(1), 108.

[28] L. A. Gatys, A. S. Ecker, M. Bethge, "A neural algorithm of artistic style," arXiv:1508.06576v2 [cs.CV] 2 Sep 2015.

[29] Z. Wang, X. Liu, L. Huang, Y. Chen, Y. Zhang, Z. Lin and R. Wang, "Model pruning based on quantified similarity of feature maps," arXiv:2105.06052v1 [cs.CV] 13 May 2021.

[30] K. Kavitha, B. T. Rao, "Evaluation of distance measures for feature based image registration using AlexNet," (IJACSA) International Journal of Advanced Computer Science and Applications, Vol. 9, No. 10, 2018.

[31] Y. Bengio, A. Courville and P. Vincent, "Representation learning: a review and new perspectives," arXiv:1206.5538v3 [cs.LG] 23 Apr 2014.

[32] H. Wang, Y. Fan, Z. Wang, L. Jiao and B. Schiele, " Parameter-free spatial attention network for person re-identification," arXiv:1811.12150v1 [cs.CV] 29 Nov 2018.

[33] S. Woo, J. Park, J. Lee and I. S. Kweon, "CBAM: convolutional block attention module," arXiv:1807.06521v2 [cs.CV] 18 Jul 2018.

[34] A. Krizhevsky, "Learning multiple layers of features from tiny images," April 8, 2009.

[35] O. Vinyals, C. Blundell, T. Lillicrap, K. Kavukcuoglu and D. Wierstra, "Matching networks for one shot learning," arXiv:1606.04080v2 [cs.LG] 29 Dec 2017.

[36] K. Song and Y. Yunhui, *NEU_surface_defect_database,* http://faculty.neu.edu.cn/yunhyan/NEU surface defect d atabase.html [08.10.2019].

[37] https://www.kaggle.com/belkhirnacim/textiledefectdetection, fabric dataset.

[38] T. Schlagenhauf, (2021): Ball Screw Drive Surface Defect Dataset for Classification. Hg. v. Karlsruher Institut für Technologie (KIT). Karlsruher Institut für Technologie (KIT) wbk Institute of Production Science. https://publikationen.bibliothek.kit.edu/1000133819.

[39] N. Mishra, M. Rohaninejad, X. Chen and P. Abbeel, "A simple neural attentive meta-learner," arXiv:1707.03141v3 [cs.AI] 25 Feb 2018.

[40] J. Hu, L. Shen, S. Albanie, G. Sun and E. Wu, "Squeeze-and-excitation networks," arXiv:1709.01507v4 [cs.CV] 16 May 2019.

[41] G. Huang, Z. Liu, L. Maaten and K. Q. Weinberger, " Densely connnected convolutional networks," arXiv:1608.06993v5 [cs.CV] 28 Jan 2018.